\documentclass[fleqn,usenatbib]{mnras}

\usepackage{newtxtext,newtxmath}

\usepackage[T1]{fontenc}

\DeclareRobustCommand{\VAN}[3]{#2}
\let\VANthebibliography\thebibliography
\def\thebibliography{\DeclareRobustCommand{\VAN}[3]{##3}\VANthebibliography}

\usepackage{graphicx}	
\usepackage{amsmath}	
\usepackage{booktabs}
\usepackage{multirow} 
\usepackage{ulem}

\newcommand{\hi}{{\rm H}{\textsc i}}

\title[Identify \hi\ Sources of CRAFTS]{Automated Identification and Segmentation of \hi\ Sources in CRAFTS Using Deep Learning Method}

\author[Song et al.]{
Zihao Song$^{1}$, 
Huaxi Chen$^{1}$\thanks{ E-mail: chenhuaxi@zhejianglab.com},
Donghui Quan$^{1}$\thanks{E-mail: donghui.quan@zhejianglab.com},
Di Li$^{2}$,
Yinghui Zheng$^{2}$,
Shulei Ni$^{1}$,
\newauthor
Yunchuan Chen$^{1}$,
Yun Zheng$^{1}$
\\
$^{1}$Zhejiang Lab, Hangzhou, Zhejiang 311121, China \\
$^{2}$National Astronomical Observatories, Chinese Academy of Sciences, Beijing 100101, People's Republic of China
}

\date{Accepted XXX. Received YYY; in original form ZZZ}

\pubyear{2015}

\begin{document}
\label{firstpage}
\pagerange{\pageref{firstpage}--\pageref{lastpage}}
\maketitle

\begin{abstract}
Identifying neutral hydrogen (\hi)  galaxies from observational data is a significant challenge in \hi\ galaxy surveys. With the advancement of observational technology, especially with the advent of large-scale telescope projects such as FAST and SKA, the significant increase in data volume presents new challenges for the efficiency and accuracy of data processing.To address this challenge, in this study, we present a machine learning-based method for extracting \hi\ sources from the three-dimensional (3D) spectral data obtained from the Commensal Radio Astronomy FAST Survey (CRAFTS). We have carefully assembled a specialized dataset, HISF, rich in \hi\ sources, specifically designed to enhance the detection process. Our model, Unet-LK, utilizes the advanced 3D-Unet segmentation architecture and employs an elongated convolution kernel to effectively capture the intricate structures of \hi\ sources. This strategy ensures a reliable identification and segmentation of \hi\ sources, achieving notable performance metrics with a recall rate of 91.6\% and an accuracy of 95.7\%. These results substantiate the robustness of our dataset and the effectiveness of our proposed network architecture in the precise identification of \hi\ sources. Our code and dataset is publicly available at \url{https://github.com/fishszh/HISF}.
\end{abstract}




\section{Introduction}

Neutral hydrogen (\hi) is a crucial constituent of the interstellar medium. Via 21 cm emission line of \hi\, researchers can study the evolution of galaxies and the distribution of matter in the universe \citep{2020A&A...638L..14C}. \hi\ emission lines provide vital information on the density and velocity structure of neutral gas within galaxies \citep{2005ApJS..160..149S}. Consequently, over the past few decades, numerous \hi\ surveys have been conducted to detect \hi\ in the local universe. Key surveys include \hi\ Parkes All Sky Survey (HIPASS) \citep{2001MNRAS.322..486B}, which identified over 5,000 galaxies across approximately 30,000 deg$^2$, and ALFALFA \citep{2005AJ....130.2598G}, which covered an area of approximately 7,000 deg$^2$, cataloging 31,502 galaxies. The FAST All Sky \hi\ Survey (FASHI) \citep{2024SCPMA..6719511Z} is a comprehensive endeavor to map the entire celestial sphere observable by the Five-hundred-meter Aperture Spherical radio Telescope (FAST), with a focus on the declination span from -14$^{\circ}$ to +66$^{\circ}$. The initial release of its data has successfully cataloged an impressive 41,741 extra-galactic \hi\ sources. In parallel, the Commensal Radio Astronomy FAST Survey (CRAFTS) \citep{2018IMMag..19..112L} leverages the same observational parameters as FAST but with an extended purview that encompasses a variety of radio astronomy targets. This includes not only \hi\ sources but also pulsars and fast radio bursts (FRBs), showcasing the versatility and depth of FAST's observational capabilities.

With the progression of observational technologies and equipment upgrades, a substantial volume of high-quality astronomical observation data has been generated through various sky survey. However, the processing of these vast datasets imposes stringent requirements on both efficiency and accuracy, which conventional methodologies struggle to fulfill. 

In response to this challenge, scientists have embarked on exploring the integration of machine learning into the data processing of astronomical observations\citep{2019arXiv190407248B}. A variety of machine learning-driven methods have been employed in diverse applications within astronomy, such as detecting tidal features \citep{2023arXiv230704967D}, light curve classification \citep{2023arXiv231108080C, Tey_2023}, source detection \citep{2023RAA....23k5006L}, spectrum classification \citep{2022arXiv220108967T}, Radio Frequency Interference (RFI) mitigation \citep{2017A&C....18...35A, 2022MNRAS.512.2025S, 2022NewA...9601825X} and so on.



In this study, we utilized CRAFTS observational data to systematically organize and construct a dedicated \hi\ sources dataset, HISF. To achieve high-precision identification and segmentation of \hi\ sources, we deployed a 3D Unet deep learning model, referred to as Unet-LK, featuring an elongated convolution kernel, which is capable of effectively extracting and segmenting \hi\ sources from complex spectral data cubes. The primary objective of this work is to enhance the accuracy and efficiency of \hi\ source detection by utilizing deep learning technology. This endeavor aims to validate the potent application potential of deep learning in astronomical data processing. Furthermore, it provides new insights and approaches for future astronomical observations and data analysis.

The paper is structured as follows: Section \ref{sec:related} introduces the related \hi\ survey and \hi\ source finding works. Section \ref{sec:data} describes the HISF dataset selection and preparation. Section \ref{sec:model} details our model pipeline and experiment results. Our summary is outlined in Section \ref{sec:summary}. 

\begin{figure*}
\begin{center}
\centerline{\includegraphics[width=.9\linewidth]{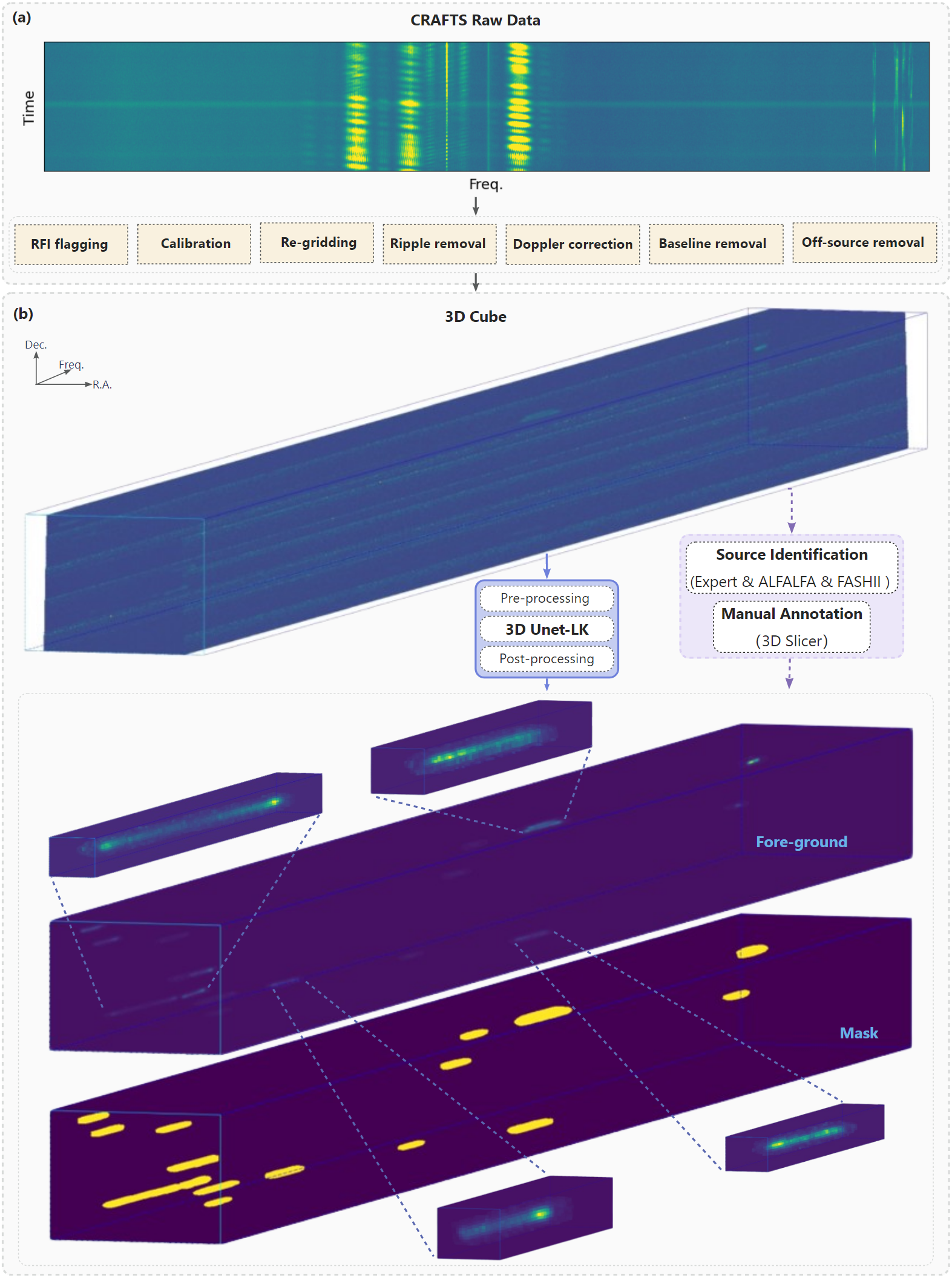}}
\caption{Data processing pipeline for \hi\ source identification. (a) The CRAFTS 3D spectral cube data in our possession is meticulously refined from the raw data. It involves critical processing steps like RFI flagging and Doppler correction, ensuring the accuracy and reliability of our observations. However, the details of these processing steps are beyond the scope of this paper. (b) Upon obtaining the 3D spectral cubes, our methodology commences with expert source identification, followed by manual annotation facilitated by 3D Slicer. Utilizing the ensuing labeled dataset, we then proceed with our model training for \hi\ source recognition.  The "Fore-ground" subplot illustrating the distribution of \hi\ source signals within the cube, including an inset that magnifies the details of these signals. The "mask" subplot delineates the regions of the annotated \hi\ sources.}
\label{fig:pipline}
\end{center}
\end{figure*}

\section{Related Work}
\label{sec:related}

In previous \hi\ surveys, for instance HIPASS, ALFALFA and FASHI, research teams conventionally developed their own automated algorithms or employed software like SoFiA\footnote{https://github.com/SoFiA-Admin/SoFiA-2} to identify \hi\ sources. These detections were subjected to further manual analysis and verification, culminating in the release of comprehensive \hi\ source catalogs. These catalogs encompassed essential attributes, including spatial coordinate ranges, frequency ranges, red-shifts, and signal-to-noise ratios (SNR), among other key parameters. Based on this spatial and frequency domain data, researchers were generally able to conduct effective analyses of the characteristics of \hi\ emission lines. 

In the realm of \hi\ source detection, extensive research efforts have been conducted. In SKA Science Data Challenge 2, several teams have devised a range of methods to identify \hi\ sources within a simulated dataset \citep{2023MNRAS.523.1967H}. These approaches not only encompass conventional methodologies like SoFiA, but also integrate machine learning techniques, such as 3D Unet for segmentation, CNN for classification, and object detection algorithms like YOLO for \hi\ source characterization. \citet{2023RAA....23k5006L} tentatively employed the Mask R-CNN model and PointRend approach to identify \hi\ signals, revealing encouraging outcomes when applied to a simulated 2D dataset. Those exploratory work subtly hints at the potential for these advanced deep learning frameworks to make a valuable contribution to the refinement and streamlining of \hi\ source detection.

Furthermore, 3D segmentation algorithms have proven to be highly applicable in the medical imaging domain, particularly with CT and MRI scans. The field of medical imaging has seen extensive application and development of these algorithms, leading to the establishment of several State-of-the-Art (SOTA) models. Notable examples include the Unet \citep{2016arXiv160606650C}, nn-Unet \citep{Isensee_2020}, FracNet \citep{ribfrac2020}, UXNet \citep{2022arXiv220915076L} and Swin-UNTR \citep{2022arXiv220101266H}, which have set benchmarks in medical image segmentation. The success of these models in delineating complex anatomical structures in 3D medical images offers valuable insights for the identification of \hi\ sources, suggesting that the principles and techniques refined in the medical context could be adapted to enhance \hi\ source detection methodologies.

\section{Data}
\label{sec:data}

Previous studies have commonly employed either conventional algorithms coupled with manual identification, or were grounded on simulated data alone, without being validated against authentic observational datasets. Furthermore, a dearth of openly accessible annotated observational datasets has hindered advancements. To address this deficiency, we utilize the observational data from CRAFTS to systematically compile a novel \hi\ source dataset, HISF, featuring accompanying masks. This innovative effort is intended to provide a much-needed benchmark for evaluating and enhancing \hi\ source identification techniques within an empirical context.

The construction of \hi\ source spectral data cubes from CRAFTS raw data follows a meticulously designed pipeline depicted in \autoref{fig:pipline}, including a series of critical steps such as RFI flagging, ripple removal, baseline removal, and other essential processing measures. Despite the inherent challenges in RFI eradication, only the most evident instances have been addressed, leaving residual RFI in the cubes. Consequently, the generated spectral data cubes still contain a substantial amount of RFI, which poses a significant challenge for identifying \hi\ sources.

\begin{figure}
\vskip 0.2in
\begin{center}
\centerline{\includegraphics[width=.98\columnwidth]{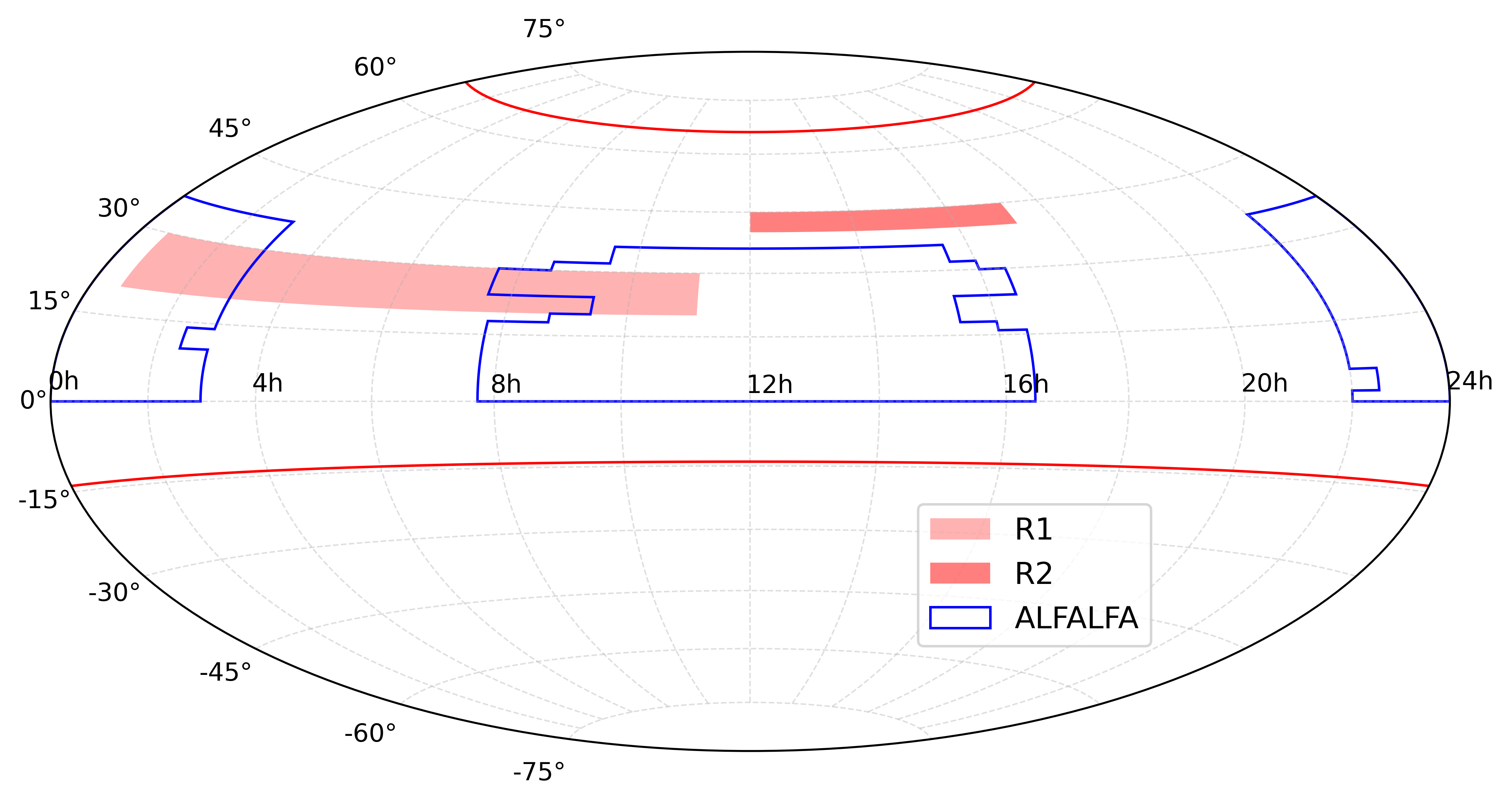}}
\caption{The region depicted between the red lines reveals the overall sky coverage of CRAFTS. Notably, the regions designated as R1 and R2 are the distinct areas where we have performed data annotation.}
\label{fig:skyCoverage}
\end{center}
\vskip -0.2in
\end{figure}

To confirm the \hi\ sources within the CRAFTS spectral data cube, we integrate expert verification and cross-validation methods utilizing other \hi\ surveys and observations in different wavebands. Currently, we have annotated data for two sky regions, as depicted in \autoref{fig:skyCoverage}.

In Region R1, we analyzed 646 3D spectral data cubes and confirmed 2050 \hi\ sources through expert verification and ALFALFA cross-validation. Among these, 1749 \hi\ sources correspond to those detected by ALFALFA. Nonetheless, they still contained unprocessed RFI signals. Due to differences in frequency coverage and sensitivity between ALFALFA and CRAFTS, we also referred to the FASHI \hi\ source catalog.

For Region R2, based on our experience accumulated during the data processing, we meticulously eliminated additional RFI signals that are commonly known to originate from electronic devices, civil aviation, and navigation satellite communications during the data processing phase. This meticulous cleaning process has resulted in a comparatively cleaner cubes. In the process of identifying \hi\ sources, we initially engaged in a manual identification of potential signals. These signals were then scrutinized through a voting system involving five experts, with a consensus of at least three affirmative votes confirming their reliability. By cross-referencing this expert validation with counterparts from other wavebands, we successfully cataloged 469 \hi\ sources.

\begin{figure}
\vskip 0.2in
\begin{center}
\centerline{\includegraphics[width=.98\columnwidth]{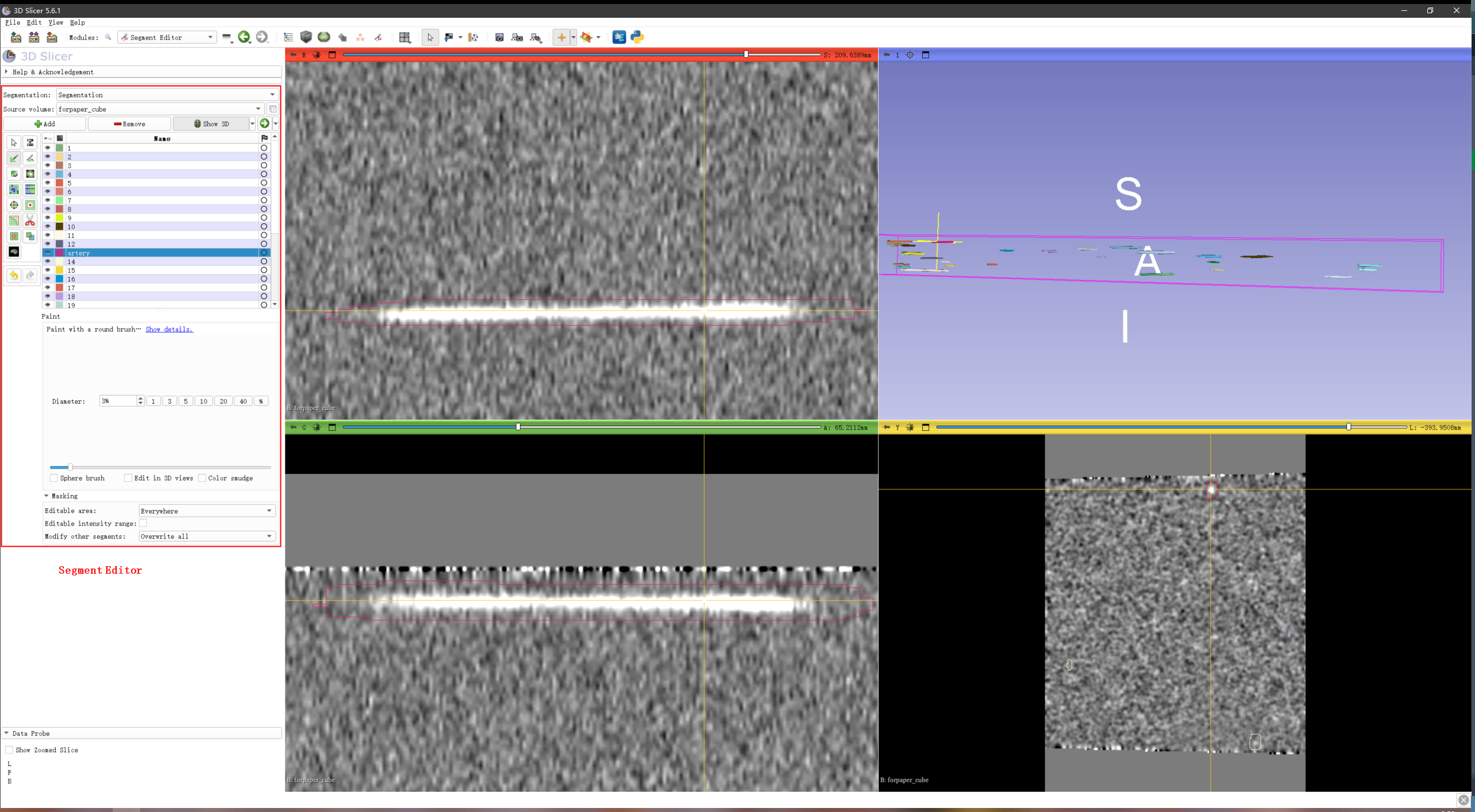}}
\caption{3D Slicer is a free, open source software package for visualization and image analysis. This is an example for visualizing 3D spectral cubes and annotating \hi\ sources. The \textbf{Segment Editor} panel is utilized for manually creating and refining (paint, draw, …) segmentations from the orthogonal planes of the 3D spectral cube. Additionally, the top right panel allows for the examination of 3D segmentations.}
\label{fig:3dslicer}
\end{center}
\vskip -0.2in
\end{figure}

Following the approximate coordinate (frequency, R.A., Dec.) of \hi\ sources from the identification process, we utilized 3D Slicer \citep{FEDOROV20121323} to visualize \hi\ signals on three orthogonal planes, see \autoref{fig:3dslicer}, and we annotate the \hi\ source on RA-Frequency plane, while checking on the other two planes. \autoref{fig:pipline} (b) displays an illustrative example of a 3D spectral cube, with the "Fore-ground" subplot showcasing the \hi\ source signals, including an inset that magnifies the details of these signals. The "Mask" subplot provides a clear visualization of the regions corresponding to the annotated \hi\ sources. The signal boundaries were not strictly defined, focusing on regions with distinct signal characteristics and a minor inclusion of non-\hi\ signal areas was tolerated. As depicted in the three orthogonal planes of \autoref{fig:3dslicer}, red contours are presented to accentuate the details of the annotated mask.

It is noteworthy to mention that, due to the difference in frequency coverage between ALFALFA and CRAFTS, despite our meticulous manual verification, the dataset might still contain a small number of instance where \hi\ sources will be either falsely identified or inadvertently omitted.

\begin{table*}
\caption{HISF dataset overview. Region R1 contains 646 spectral cubes and 2050 \hi\ sources, with a spatial resolution of 0.0167 degree/pixel and a frequency resolution of 7.6 kHz/pixel. Each cube spans 3930 to 3932 pixels in the frequency direction, encompasses 23 pixels in the R.A. direction, and ranges from 231 to 261 pixel in the DEC. direction. Region R2 includes 157 spectral cubes and 469 \hi\ sources, maintaining the same spatial and frequency resolutions. Each cube in region R2 extends from 3275 to 4325 pixels in the frequency direction, varies from 158 to 181 pixels in the R.A. direction, and spans from 191 to 248 pixel in the DEC. direction.}
\label{tab:sources}
\vskip 0.15in
\begin{center}
\begin{small}
  \begin{tabular}{cccccc}
    \toprule
    \multirow{2}{*}{Region} & \multicolumn{3}{c}{No. Cube}     & \multirow{2}{*}{No. Source}  & \multirow{2}{*}{Shape (pixel)}  \\  
                \cline{2-4}      \\                        
                            & Train  & Valid  & Test                \\
    \midrule \\
    R1                      & 540     & 36     & 70        & 2050    & (3930-3932,23,231-261)  \\
    R2                      & 100     & 15     & 42        & 469     & (3275-4325,158-181,191-248) \\
    \bottomrule
  \end{tabular}
\end{small}
\end{center}
\vskip -0.1in
\end{table*}

\begin{figure}
\vskip 0.2in
\begin{center}
\centerline{\includegraphics[width=.95\columnwidth]{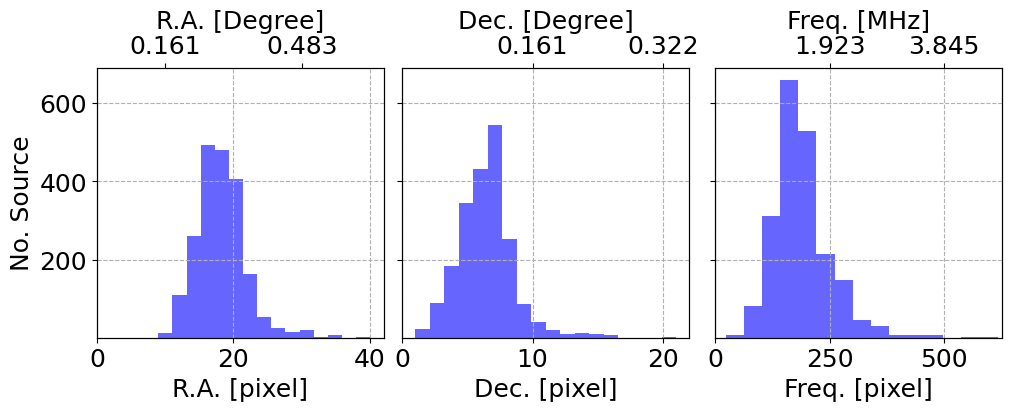}}
\caption{The distribution of \hi\ source extents across R.A., DEC. and frequency axes, measured in pixel units, with a spatial resolution of 0.0167 degrees/pixel and a frequency resolution of 7.6 kHz/pixel. The \hi\ sources exhibit a pronounced elongation, with their frequency pixel span markedly exceeding the spatial dimensions.}
\label{fig:statistic}
\end{center}
\vskip -0.2in
\end{figure}

We conducted a basic statistical analysis on the source size of all \hi\ sources, see \autoref{fig:statistic}. From the figure, it is evident that the \hi\ sources occupy a significantly greater number of pixels along the frequency axis than in the spatial dimensions, thus exhibiting an elongated shape. 


The HISF dataset comprises 3D spectral cubes from Regions R1 and R2. For Region R1, the dataset is partitioned into 540 cubes for training, 36 for validation, and 70 for testing. Region R2 contributes 100 cubes to the training set, 15 to the validation set, and 42 to the test set, see \autoref{tab:sources}.

In accordance with the data release policy of FAST, the data from Region R1 is anticipated to be made publicly available on HIverse\footnote{https://hiverse.alkaidos.cn/} in the near future, while the data from Region R2 will be released at a later date. HISF dataset is formatted as pairs consisting of 3D spectral cubes and their corresponding labels.

\section{Model and Experiments}
\label{sec:model}
\begin{figure}
\vskip 0.2in
\begin{center}
\centerline{\includegraphics[width=.98\columnwidth]{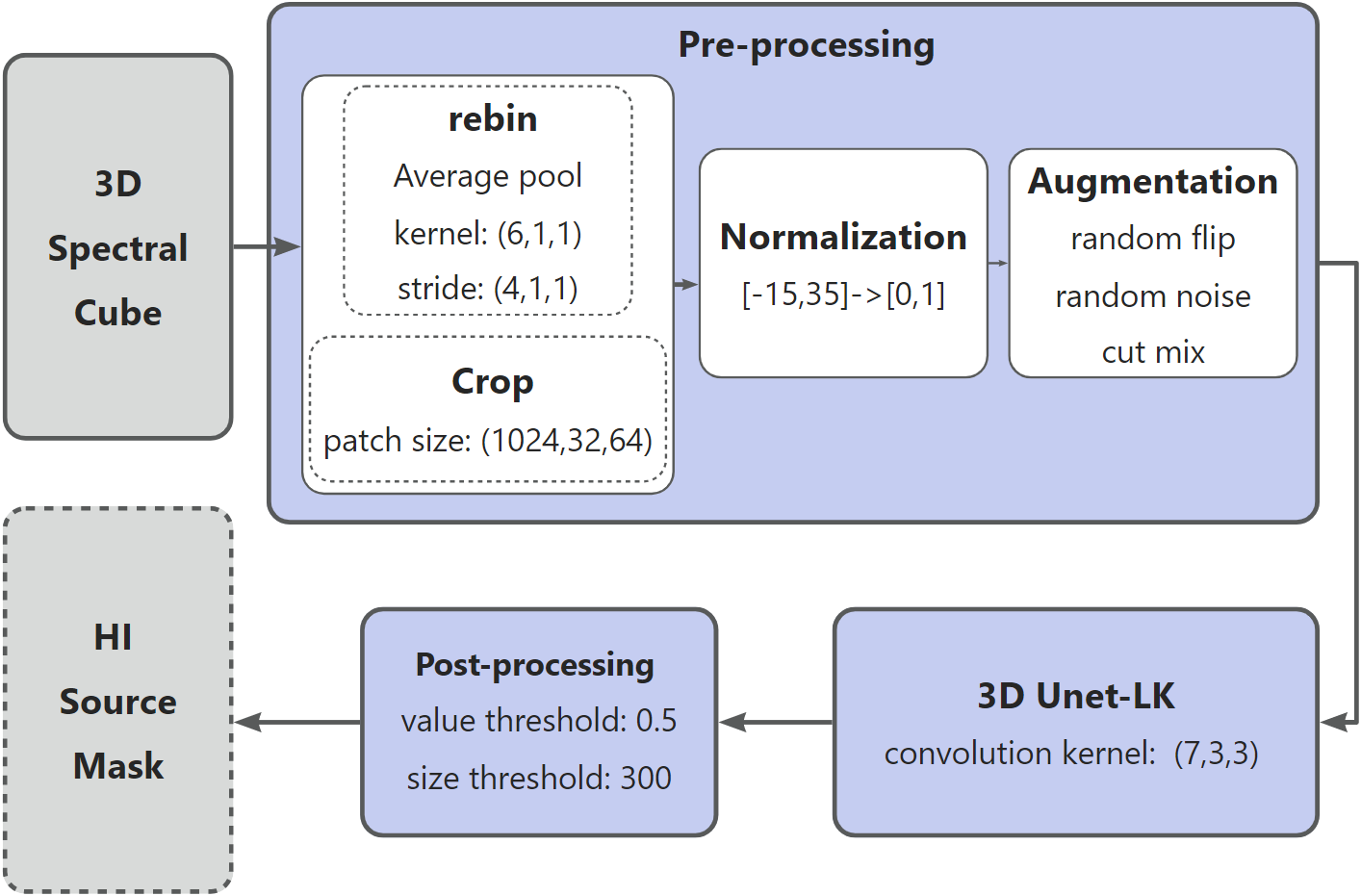}}
\caption{The model pipeline for identifying \hi\ sources, including data pre-processing, model training, and post-processing to refine the results. Two strategies, \textbf{rebin} and \textbf{crop}, can be applied either individually or in combination, as shown in \autoref{tab:performance}.}
\label{fig:modelpipeline}
\end{center}
\vskip -0.2in
\end{figure}

In the CRAFTS spectral data cube, identifying \hi\ sources is regarded as a target detection problem within deep learning. Given Unet's outstanding performance in 3D image segmentation and object detection tasks\citep{2016arXiv160606650C, ribfrac2020, Isensee_2020}, we employ a 3D-Unet architecture as the fundamental framework for this task, as its precise segmentation facilitates subsequent \hi\ emission line analysis. As illustrated in \autoref{fig:modelpipeline}, our model pipeline consists of three stages: (a) pre-processing, (b) model training, (c) post-processing.

\subsection{Model pipeline}
\textbf{(a) Pre-processing:} Since a whole-volume 3D spectral cube could be too large to fit in a regular GPU memory, we implement two strategies, (1) \textbf{rebin}: apply an average pooling layer with convolution kernel of (6,1,1) and stride of (4,1,1) on the frequency axis to reduce its dimensionality, (2) \textbf{crop}: conduct segmentation in a sliding window manner, adopting a patch size of (1024, 32, 64) with a stride equal to half the patch size. These two strategies permit an increased batch size, thereby enhancing the efficiency of the training process.

The intensity of input cubes were clipped to the window [-15, 35], then normalized to the range [0, 1].

Simultaneously, for data augmentation, we employ random flipping across the R.A., Dec. and frequency axes with a probability 0.5 along each axis. We also add random gaussian noise, with $\mu=0$ and $\sigma \sim U(2.8, 3.8)$, statistically derived from the 3D spectral cube.

To improve the recognition capability of faint \hi\ signals, we employed the cut mix technique, randomly degrade the intensity of high SNR \hi\ sources, mimicking weaker \hi\ signals by extracting them from the labeled mask and adjusting their intensity to 30-80\% of the original. This technique is employed to augment our dataset with a variety of signal strengths.

\begin{figure}
\vskip 0.2in
\begin{center}
\centerline{\includegraphics[width=.98\columnwidth]{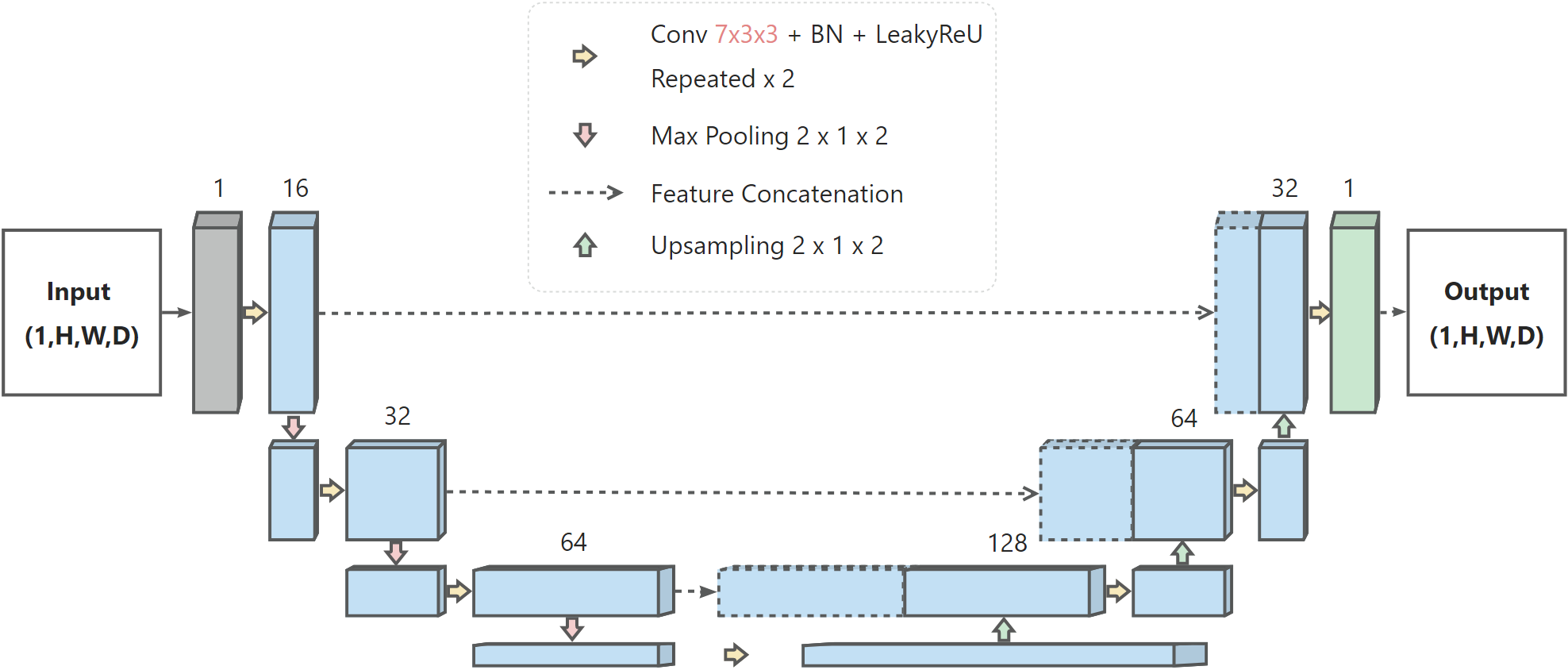}}
\caption{The neural network architecture of UNet-LK is based on a four-layer 3D UNet, featuring an elongated convolution kernel of size (7, 3, 3) along the frequency axis.}
\label{fig:unet}
\end{center}
\vskip -0.2in
\end{figure}

\textbf{(b)  Model training:} Given that the frequency resolution of \hi\ sources in CRAFTS spectral cube data is significantly higher than the spatial resolution, they span approximately ten pixels in space but encompass hundreds of pixels along the frequency axis, see \autoref{fig:statistic}. To address this disparity, we have employed a four-layer 3D Unet network, \autoref{fig:unet}, named Unet-LK, utilizing an elongated convolution kernel of size (7,3,3) along the frequency axis to capture additional contextual information.

Throughout the training process, we utilize the Adam optimizer with an exponentially decaying learning rate from 0.01 to 0.0005, and combine dice loss ($L_{Dice}$) and binary cross-entropy loss ($L_{Cross}$) functions (\autoref{eq:loss}) to train the network. The model is trained for a total of 600 epochs with a batch size of 2, utilizing an NVIDIA A40 GPU.

\begin{equation}
    Loss = L_{Dice} +  0.5 L_{Cross}
	\label{eq:loss}
\end{equation}

\begin{equation}
    Dice = \frac{2}{N} \frac{\sum_{i}^N y_{i} \cdot \hat{y}_{i} + \epsilon}{\sum_{i}^N y_{i} + \sum_{i}^N \hat{y}_{i} + \epsilon}
    \label{eq:dice}
\end{equation}

\begin{equation}
    L_{Dice} = 1- Dice
\end{equation}

where N is the number of spectral cube, $y_i$ is the masked label, and $\hat{y}_i$ is the model prediction output.

\textbf{(c) Post-processing} To generate the segmentation outcomes, for areas of overlap, we take the average value, and we binarize the post-processed outcomes using a threshold of 0.5. To efficiently reduce the false positive in our predictions, predictions of small sizes (smaller than 300 voxels) were filtered out.

\subsection{Experiments}
To enhance the comprehensiveness and depth of our comparative analysis, we also utilized SoFiA on HISF dataset, with an SoFiA setup: detection threshold is 5$\sigma$; smoothing kernels are kernelsXY = 0, 3, 6 and kernelsZ = 0, 3, 7, 15; the minimum number of spatial and spectral pixels is 5 in XY and Z space, while the maximum size is 50 pixels in XY space, but not limited in Z space.

In addition, we have also employed two  state-of-the-art (SOTA) frameworks, namely Swin-UNETR and UX-Net, on the 3D medical image segmentation task. Both native and re-bined resolutions were considered for the input volume data. We maintain a balanced ratio of 1:1 for positive and negative samples, ensuring each class is adequately represented during training. Specifically, positive samples are cropped in a manner that ensures they encapsulate at least half the area of the \hi\ source, whereas negative samples are randomly extracted within the confines of the spectral data cube. This strategy allows the model to learn more effectively from the target areas and enhances its segmentation performance.

\begin{figure*}
\begin{center}
\centerline{\includegraphics[width=.95\linewidth]{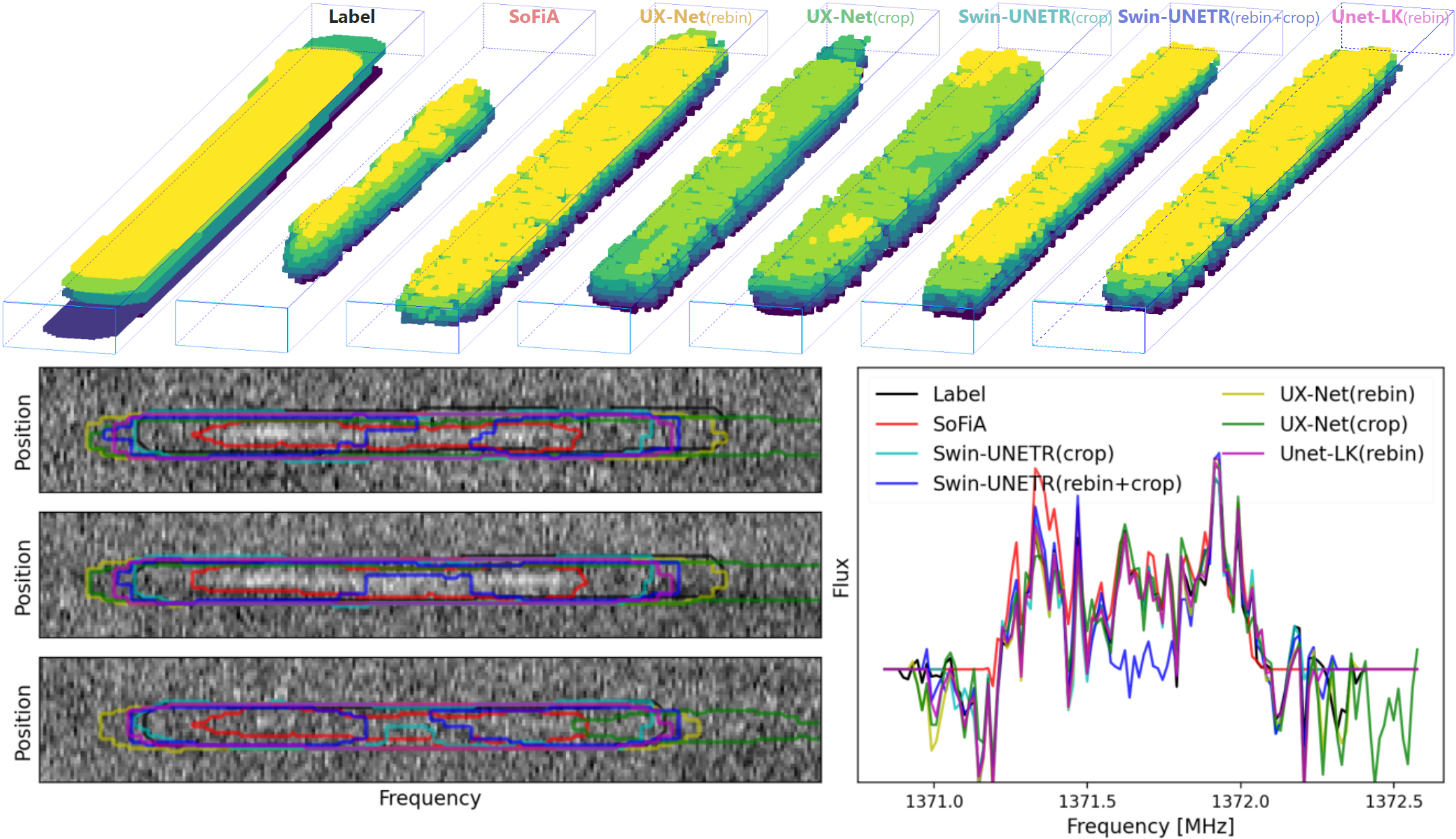}}
\caption{A visualization of the predicted segmentation for one \hi\ source from test dataset. The top panels display the segmentation mask results comparison of all methods in 3D view. The bottom left panels show the segmentation details on three frequency-position slices. The bottom right panel presents a comparison of the smoothed \hi\ emission lines. The presence of negative flux in the spectra is primarily attributed to the insufficient baseline removal during the batch processing phase, \autoref{fig:performance2} illustrates three examples that are free from this issue.}
\label{fig:performance1}
\end{center}
\end{figure*}

\begin{figure*}
\begin{center}
\centerline{\includegraphics[width=.95\linewidth]{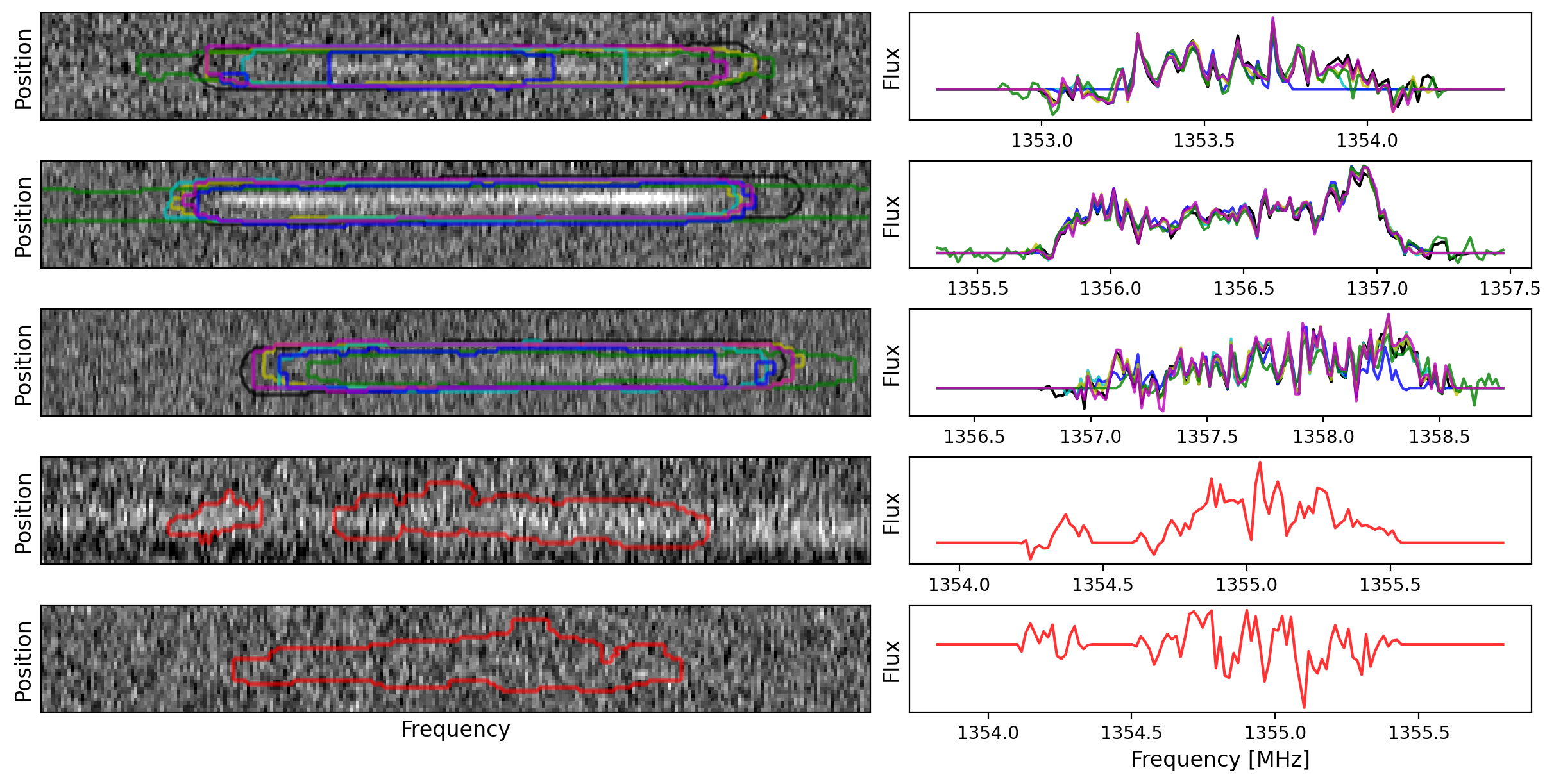}}
\caption{Same as \autoref{fig:performance1}, these are three additional segmentation examples where SoFiA failed to detect all three \hi\ sources, and two SoFiA false positive examples.}
\label{fig:performance2}
\end{center}
\end{figure*}

\subsection{Results}
We employ a segmentation algorithm to accomplish the detection task, hence we utilize both Dice (\autoref{eq:dice}) and Intersection over Union (IoU, \autoref{eq:iou}) metrics to evaluate the model's performance. Given the high threshold configuration within SoFiA, which predominantly identify \hi\ sources with high SNRs, we have adopted an IoU threshold of greater than or equal to 0.2 as the criterion for successful detection.

\begin{equation}
    IoU = \frac{Area\ of\ Intersection}{Area\ of\ Union}
	\label{eq:iou}
\end{equation}

As illustrated in \autoref{fig:performance1} and \autoref{tab:performance}, our method successfully attains a recall rate of 91.6\% and an impressive accuracy rate of 95.7\%, which distinctly surpasses the performance of the commonly employed SoFiA. Notably, our approach demonstrates exceptional capability not only in recognition precision but also achieves an acceptable level of segmentation effectiveness. The dice coefficients for our method reach 78.4\% on the training set, 74.3\% on the validation set, and 72.6\% on the test set. 

Relative to the performances achieved by Swin-UNETR and UX-Net, our proposed method displays enhanced results in both recall and precision metrics, possibly due to the fact that the elongated morphology of \hi\ sources necessitates a larger global receptive field. By addressing this need, our approach seems to be more adept at handling such structural intricacies, leading to improved recognition and segmentation outcomes.These results firmly substantiate the stability and generalization capabilities of our network, exhibiting formidable strengths in both precise source identification and effective data segmentation. 

It is worth noting that the preliminary data processing, from raw data to cube data, significantly impacts the detection performance of SoFiA, with RFI being a primary cause of low precision. Since SoFiA can only identify sources with 
positive values, inadequate baseline removal can greatly affect its recall rate. Nonetheless, deep learning models are equipped to mitigate these issues to a considerable extent.

\begin{table*}
\caption{A performance comparison of SoFiA, Unet-LK, Swin-UNETR and UX-Net on test set, with 368 \hi\ sources. Due to limitations in GPU memory, both \textbf{crop} and \textbf{rebin} techniques were employed during the pre-processing phase for Unet-LK, UX-Net, and Swin-UNETR. For the high threshold configuration within SoFiA, in this study, we adopt IoU (\autoref{eq:iou}) $\ge$ 0.2 as the detection threshold criterion.}
\label{tab:performance}
\begin{center}
\begin{small}
  \begin{tabular}{cccccccccc}
    \toprule
    Methods                    & Param  & \multicolumn{2}{c}{Segmentation}    & & \multicolumn{4}{c}{Detection}   \\  
                                        \cline{3-4}                           \cline{6-9}       
                               &        & IoU   & Dice                        & & Recall    & Precision  & TP (Number) & FP (Number)\\
    \midrule 
    SofiA                      &        &  1.5\%  &  2.9\%                    & & 64.2\%    &  2.3\%     & 236  & 10036\\
    UX-Net (rebin)             &  22.5M & 56.0\%  & 71.8\%                    & & 89.4\%    & 93.5\%     & 329  & 23  \\
    UX-Net (crop)              &  22.5M & 49.7\%  & 66.4\%                    & & 89.7\%    & 78.9\%     & 330  & 88  \\
    Swin-UNETR (crop)          &  15.5M & 45.6\%  & 62.7\%                    & & 90.5\%    & 51.2\%     & 333  & 317 \\
    Swin-UNETR (rebin+crop)    &  15.5M & 47.8\%  & 64.7\%                    & & 85.9\%    & 90.3\%     & 316  & 34  \\
    \textbf{Unet-LK} (rebin)   &  7.2M  & \textbf{59.1\%}  & \textbf{74.3\%}  & & \textbf{91.6\%}  & \textbf{95.7\%}  & \textbf{337} & \textbf{15} \\
    \bottomrule
  \end{tabular}
\end{small}
\end{center}
\vskip -0.1in
\end{table*}

\section{Summary}
\label{sec:summary}
In this work, we propose a novel method for \hi\ source detection that harnesses the power of 3D-Unet segmentation network to accurately identify and segment \hi\ sources. Experimental results demonstrate remarkable performance on our test set, achieving high recall (91.6\%) and accuracy (95.7\%), while maintaining good consistency across different datasets.

Compared to the SoFiA software, our proposed method demonstrates a significant improvement in recognition precision and attains satisfactory segmentation outcomes within the context of HISF dataset. It exhibits superior generalization capabilities, effectively mitigating the impact of RFI and other data processing artifacts to a certain extent. Comparative analysis with state-of-the-art network architectures such as Swin-UNETR and UX-Net indicates that customizing the network architecture in accordance with the specific attributes of the data and target features is indeed a critical factor in optimizing the overall functionality and performance of the model. This not only validates the efficacy of our adopted method but also highlights the profound value of our tailored HISF dataset in enhancing the precision and efficiency of \hi\ source detection tasks.

Additionally, the meticulously constructed and annotated custom HISF dataset we have developed plays a pivotal role in future identification tasks concerning \hi\ sources. HISF dataset offers a comprehensive collection of \hi\ source instances, covering a wide range of observing conditions and signal strengths, with a particular emphasis on cases where \hi\ sources are difficult to discern amidst complex background noise and low SNR environments. Our meticulous manual annotation process guarantees the authenticity and completeness of each source within the HISF dataset, a critical factor for the training and validation of \hi\ source identification algorithms.

Despite its promising performance, the proposed method has potential for refinement. Improving the model's sensitivity to low SNR \hi\ sources is a notable aspect. Additionally, noise and data variability in \hi\ datasets might affect generalizability across diverse environments. Future work could thus focus on refining pre-processing techniques to handle these complexities and enhancing network resilience to SNR variations.

Furthermore, given the success with our custom HISF dataset and architecture, future directions include expanding the dataset diversity, developing adaptive learning strategies, and exploring ways to integrate extra contextual information to boost the accuracy of \hi\ source identification and segmentation.

In conclusion, the promising outcomes of this research have not only made a substantial contribution to the advancement of \hi\ source detection methodologies, but also revealed an expanded scope of potential applications within the critical task of extracting and analyzing complex sources in the realms of radio astronomy and its associated domains.

\section*{Acknowledgements}

This work was Supported by National Key R$\&$D Program of China No. 2023YFE0110500 and No. 2022YFB4501405, National Natural Science Foundation of China grant No. 12373026 and the Leading Innovation and Entrepreneurship Team of Zhejiang Province of China Grant No. 2023R01008.

\section*{Data Availability}

The labeled CRAFTS Data used in this paper will be available in the near future, and the data access URLs will be synchronized on HIverse and GitHub  \url{https://github.com/fishszh/HISF}.



\bibliographystyle{mnras}
\bibliography{reference} 








\bsp	
\label{lastpage}
\end{document}